\documentclass{article} % For LaTeX2e
\usepackage{iclr2024_preprint, times} % Used for arxiv
\iclrfinalcopy
\usepackage[utf8]{inputenc} % allow utf-8 input
\usepackage[T1]{fontenc}    % use 8-bit T1 fonts
\usepackage{hyperref}       % hyperlinks
\hypersetup{
    colorlinks,
    linkcolor={red!50!black},
    citecolor={blue!50!black},
    urlcolor={blue!80!black}
}
\usepackage{color}
\definecolor{blue}{HTML}{1f77b4}
\definecolor{red}{HTML}{d62728}
\definecolor{olivie}{HTML}{bcbd22}
\definecolor{cyan}{HTML}{17becf}
\usepackage{spverbatim}
\usepackage{enumitem}
\usepackage{url}            % simple URL typesetting
\usepackage{booktabs}       % professional-quality tables
\usepackage{amsfonts,amsmath,amssymb,mathtools}       % blackboard math symbols
\usepackage{nicefrac}       % compact symbols for 1/2, etc.
\usepackage{microtype}      % microtypography
\usepackage{bm}
\usepackage{multirow}
\usepackage{array}
\usepackage{tabularx}
\usepackage{algorithm,algpseudocode}
\usepackage{setspace}
\usepackage{wrapfig}
\usepackage{multicol}
\usepackage{bbm}
\usepackage{subcaption}
\usepackage{listings}
\usepackage{xspace}
\usepackage{verbatimbox}
\usepackage{glossaries}
 
\usepackage{cleveref}
\usepackage{siunitx}

\algrenewcommand\algorithmicrequire{\textbf{Input:}}
\algrenewcommand\algorithmicensure{\textbf{Output:}}
\algnewcommand{\LineComment}[1]{\State // \textit{#1}}
\newcommand{\parhead}[1]{\textbf{#1}~}

\newcommand{\LAMA}{LLaMA-13b\xspace}

\newcommand{\x}{\mathbf{x}}

\newcommand{\X}{\mathbf{X}}

\newcommand{\DW}{\Delta{W}}
\newcommand{\R}{\mathbb{R}}

\newcommand{\W}{\mathbf{W}}
\newcommand{\y}{\mathbf{y}}

\DeclarePairedDelimiterX{\infdivx}[2]{(}{)}{%
  #1\;\delimsize|\delimsize|\;#2%
}

\makeglossaries
\newglossaryentry{llm}
{
    name=LLM,
    description={large language model}
}

\newglossaryentry{nn}
{
    name=NN,
    description={neural network}
}

\newglossaryentry{ood}
{
    name=OOD,
    description={out-of-distribution}
}

\newglossaryentry{lora}
{
    name=LoRA,
    description={low-rank adapter}
}

\newglossaryentry{mqa}
{
    name=multiple-choice QA,
    description={multiple-choice QA}
}

\newglossaryentry{mlm}
{
    name=MLM,
    description={masked language model}
}

\newcommand{\EOL}{LoRA ensembles\xspace}

\title{LoRA ensembles for large language model fine-tuning}
\author{%
  Xi Wang\thanks{Work done during an internship at the Bosch Center for AI. Contact: \texttt{xwang3@cs.umass.edu}} \\
  UMass Amherst\\
  \And
  Laurence Aitchison\\
  University of Bristol\\
  \And
  Maja Rudolph\\
  Bosch Center for AI \\
}

\begin{document}
\maketitle

% \begin{abstract}
%     Large language models (LLMs) are pre-trained on massive text corpora and exhibit impressive emergent capabilities. Still, to get improvement gains for specialized tasks (in this paper we study question answering (QA) on specialized domains) LLMs benefit from fine-tuning. We find that fine tuning leads to poor calibration and that deep ensembles are not efficient or accurate enough to overcome calibration issues in LLM fine tuning. In our extensive empirical study on fine-tuning LLAMA 13b on various QA tasks, we find that two ingredients are essential for improving calibration: ensembling needs to happen on the level of lowrank components and each of the components needs to be regularized strongly.
%     Based on these findings we propose efficient ensemble of regularized models (EERM). With EERM we achieve on average +0.x on a battery of QA taks.
% \end{abstract}
 \vspace{-1pt}
\begin{abstract}
 \vspace{-1pt}
Fine-tuned LLMs often exhibit poor uncertainty quantification, manifesting as overconfidence, poor calibration, and unreliable prediction results on test data or out-of-distribution samples. 
One approach commonly used in vision for alleviating this issue is a deep ensemble, which constructs an ensemble by training the same model multiple times using different random initializations.
% While each single model still shows overconfidence prediction, the averaged predictve distribution usually demonstrates improved calibration and good uncertainty quantification.
However, there is a huge challenge to ensembling LLMs: the most effective LLMs are very, very large.
Keeping a single LLM in memory is already challenging enough: keeping an ensemble of e.g. 5 LLMs in memory is impossible in many settings.
%the application of deep ensembles to the fine-tuning of large language models (LLMs) presents two distinct challenges: standard fine-tuning initializes with pre-trained models, hindering random initialization, and maintaining multiple LLM instances is practically expensive. 
% However, LoRA allows...
To address this issue, we propose an ensemble approach using Low-Rank Adapters (LoRA), a parameter-efficient fine-tuning technique. 
% The adapters' random initialization allows for model diversity, while its low-rank property minimizes storage and computational costs.
Critically, these low-rank adapters require a very small number of parameters, orders of magnitude less than the underlying pre-trained model. 
Thus, it is possible to construct large ensembles of LoRA adapters with almost the same computational overhead as using the original model.
We find that \EOL, applied on its own or on top of pre-existing regularization techniques, gives consistent improvements in predictive accuracy and uncertainty quantification.
%We found that such a  regularized ensemble gave accurate and calibrated predictions.
%Next, we combined \EOL with pre-existing regularization techniques.
%We found that \EOL gave
%\EOL gives further improvements of pre-existing regularization techinques, implying that  give fu improvements , our experiments show that ensemble only partially mitigates overconfidence in small datasets and we notice that by further incorporating regularization techniques together with ensemble, we can achieve both accurate and calibrated predictions.
\end{abstract}
% \iffalse
% \begin{abstract}
% Modern deep learning models usually demonstrate poor uncertainty quantification ability, it often shows overconfidence, badly calibrated and unreliable prediction results on data unseen during training time. In order to alleviate this issue, a very common approach is to use deep ensemble, which trains the same model multiple times to construct an ensemble when making predictions. While each single model still shows overconfidence prediction, the averaged predictve distribution usually demonstrates improved calibration and good uncertainty quantification.
% In this paper, we show that deep ensemble is not sufficient for fighting overconfidence in the large language model (LLM) fine-tuning, instead, one can simply fine-tune with extremely large weight decay, which achieves performance similar to deep ensemble in terms of calibration and uncertainty quantification while only taking a single forward pass. 
% \end{abstract}
% \fi
\section{Introduction}
% TODO: Add dropout regularization and ensemble
% TODO: Add KL trace to the appendix.

\begin{figure}[t]

\begin{subfigure}[t]{0.42\linewidth}
 \centering
    \includegraphics[width=\linewidth]{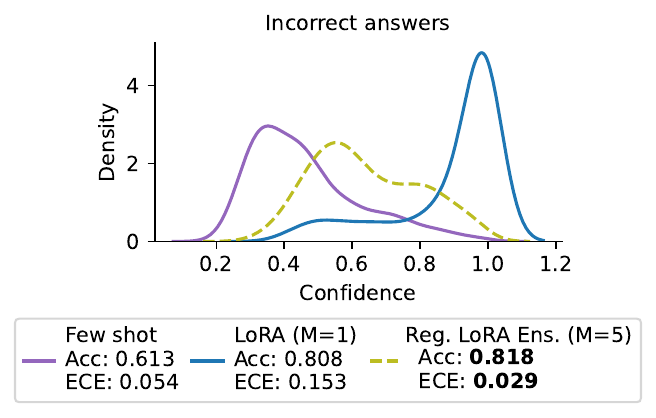}
    % \xw{Change it to regular question}}
    \caption{}
    \label{fig:overconfidence_viz}
\end{subfigure}
\begin{subfigure}[t]{0.57\linewidth}
 % \centering
    % \raisebox{8mm}{\includegraphics[width=\linewidth]{figs/prompt_template.pdf}}
    % {\includegraphics[width=\linewidth]{figs/prompt_example.pdf}}
    \raisebox{2mm}{\includegraphics[width=\linewidth]{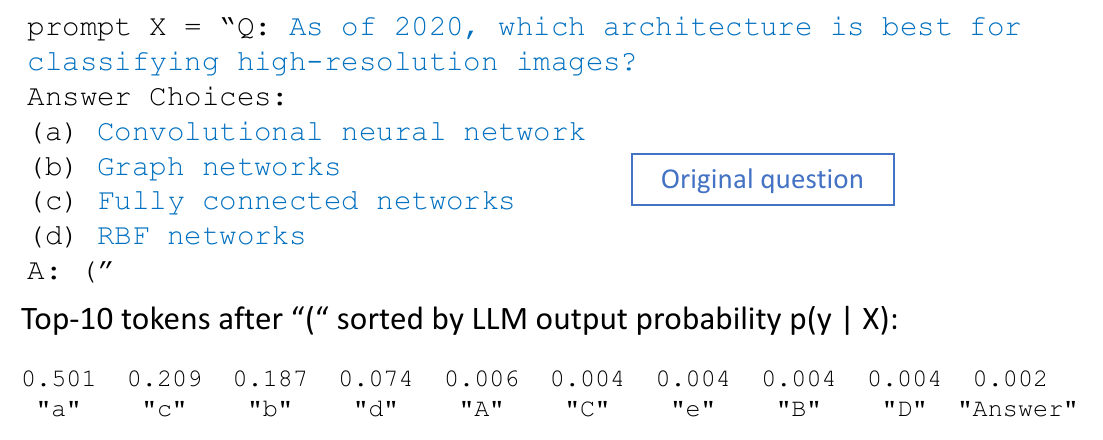}}
    % \xw{Change it to regular question}}
    \caption{}
    \label{fig:prompt_template}
\end{subfigure}
\vspace{-3pt}
\caption{%Our proposed method LoRA Ens. macht
\textbf{LoRA ensembles with strong weight decay regularization, is more accurate and better calibrated than fine-tuning a single LoRA component on multiple-choice QA problems such as in Fig.~\ref{fig:prompt_template}.} %from the cqa dataset. 
%fine-tuned LLM using LoRA shows high confidence (measured by maximum softmax probability) when making wrong predictions (\ref{fig:overconfidence_viz}), resulting in high calibration error.
Fig~\ref{fig:overconfidence_viz}, shows a KDE of the confidence with which a pre-trained \LAMA in the few-shot setting (purple line), a 
 fine-tuned LoRA model (blue line), and our proposed \EOL (yellow dashed line) make wrong predictions on the cqa dataset. 
The few-shot approach is well-calibrated but often wrong, while LoRA (M=1) is more accurate but overconfident in its wrong predictions. Our approach provides improvements in both accuracy and calibration in terms of ECE.
 }
 \vspace{-5pt}
%\caption{\textbf{On multiple-choice QA problems (\ref{fig:prompt_template}), fine-tuned LLM using LoRA shows high confidence (measured by maximum softmax probability) when making wrong predictions (\ref{fig:overconfidence_viz}), resulting in high calibration error.} On the MMLU social sciences dataset, the \emph{pre-trained} \LAMA in a 3-shot setting is less accurate yet well-calibrated, and the \emph{fine-tuned} version (blue line) boasts higher accuracy at the cost of increased expected calibration error. Our proposed \emph{LoRA Ensembles} with regularization (chosen as weight decay in the figure) achieves both accuracy and calibration.}
\end{figure}

LLMs have demonstrated state-of-art performance in many natural language processing tasks~\citep{radford2019language,touvron2023llama,brown2020language,chung2022scaling,kojima2022large,OpenAI2023GPT4TR}. 
With additional fine-tuning a pre-trained LLM can be adapted to  downstream applications or data.
However, fine-tuned LLMs can overfit to training data and often exhibit \emph{overconfidence} (as visualized in Fig.\ref{fig:overconfidence_viz}). 
Specifically, these models may yield overly certain predictions, especially on incorrectly predicted samples or those from different domains. 
Ideally, a model should exhibit low confidence when its predictions are likely to be incorrect; otherwise, the outcomes could be dangerously misleading in safety-critical contexts such as medical diagnosis\citep{singhal2023large}, finance~\citep{yang2023fingpt}, or decision-making processes~\citep{li2022pre}.

A widely adopted approach for mitigating overconfidence in deep learning is to make predictions using an \emph{ensemble} of neural networks rather than a single model. 
There are many approaches for constructing an ensemble of networks, such as training multiple networks with different random initializations~\citep{lakshminarayanan2017simple}, different hyperparameters~\citep{wenzel2020hyperparameter}.
% , or collecting the model checkpoints throughout the optimization or sampling trajectory~\citep{huang2017snapshot, Zhang2020Cyclical}. 
However, there are two barriers to applying these approaches for fine-tuning LLMs.
First, ensembles require storing multiple copies of the model weights and loading them into GPU at test time. 
This is not practical for modern LLMs. 
A single \LAMA~\citep{touvron2023llama} stored at 16-bit precision, is \qty{25}{\giga\byte} on disk, and loading it to the GPU takes around 6 seconds.  
In addition, random initialization has been noted to play a crucial role in deep ensembles~\citep{lakshminarayanan2017simple, fort2019deep}. 
However, starting the fine-tuning of the individual LLMs with the same initialization -- the pre-trained weights -- eliminates an important source of randomness and may cause a lack of diversity across the ensemble, thereby potentially reducing its benefits.

% , while loading a low-rank adapter only takes 30Mb and 0.1 seconds for loading

Work by \citet{gleave2022uncertainty} and \citet{sun2022quantifying} has attempted building ensembles of fine-tuned LLMs but due to the limitations above, their method is restricted to smaller models such as GPT-2 \citep{radford2019language} with only 1.5 billion parameters. In this paper, we build on recent advances in efficient LLM fine-tuning with low-rank adapters (LoRA)~\citep{hu2021lora} and propose an ensemble method for LLM fine-tuning that scales to models with 13 billion parameters and beyond. We propose \EOL. 
%Our method is built upon low-rank adapter (LoRA)~\citep{hu2021lora}, a parameter efficient method for LLM fine-tuning. 
\EOL solves the two aforementioned issues: LoRA requires orders of magnitude less storage than the original model: a low-rank adapter for \LAMA is only 30Mb on disk and takes 0.1 seconds to load onto GPU. 
In addition, the random initialization of the adapter provides the necessary randomness for variability across the ensemble components.

Our empirical results on several commonsense reasoning tasks show that \EOL improve accuracy and calibration over naive LoRA fine-tuning and produces better ensembles than alternatives based on last-layer fine-tuning ~\citep{du2021fewshot} or Monte Carlo dropout ~\citep{gal2016dropout}. 
% Additionally, we show that \EOL can be applied on top of regularization methods, which we call \eorl. The regularization in \eorl gives additional benefits, predominantly in calibration though also to some extent in accuracy.
As an additional contribution we study regularized \EOL. Classical theory~\citep{breiman2001random} suggests that the generalization performance of ensembling depends on the diversity of individual components. While no comparable results exist for neural networks it is believed that this intuition still holds for deep ensembles~\citep{lakshminarayanan2017simple, fort2019deep}. Initialization of the LoRA ensemble components around the same pre-trained weights already introduces a strong correlation between the ensemble components and regularization can further strengthen this effect. Yet we find in an extensive empirical study of LoRA ensembles in combination with different regularization strategies that LoRA ensembles are compatible with regularization and their combination typically further improves prediction and calibration accuracy.

\begin{figure}[t]
        \centering
    \begin{subfigure}[t]{\linewidth}
    \centering
     \includegraphics[width=\linewidth]{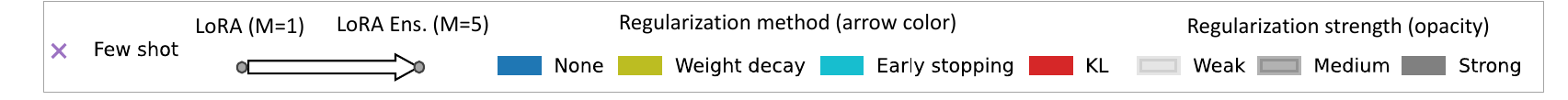}
    \caption*{}
    \end{subfigure} \\[-3.8ex]

    \centering
    \begin{subfigure}[b]{0.325\linewidth}
    \includegraphics[width=\linewidth]{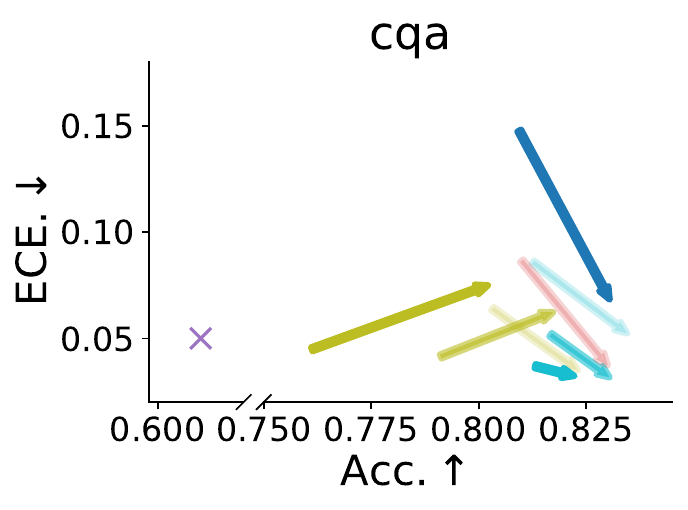}
    \end{subfigure}
    \begin{subfigure}[b]{0.325\linewidth}
   \includegraphics[width=\linewidth]{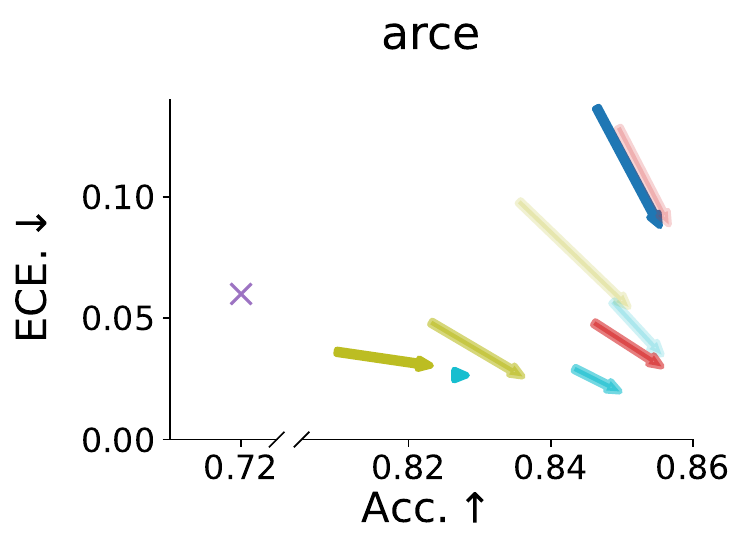}
    \end{subfigure}
    \begin{subfigure}[b]{0.325\linewidth}
   \includegraphics[width=\linewidth]{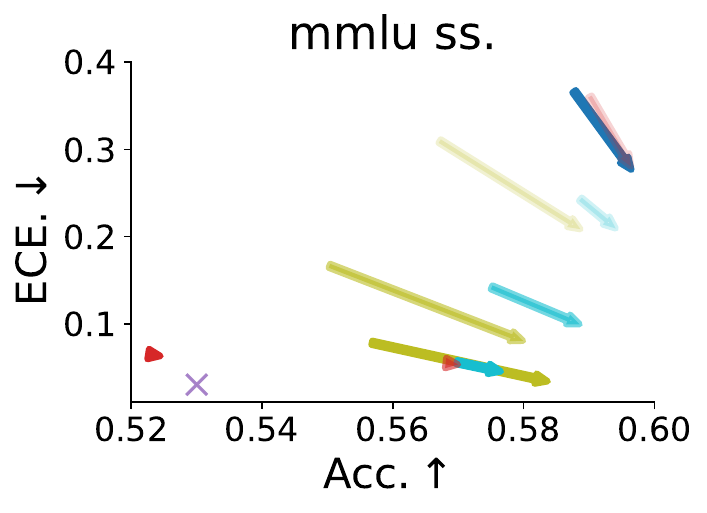}
    \end{subfigure}
    \vspace{-2pt}
    \begin{subfigure}[b]{0.325\linewidth}
    \includegraphics[width=\linewidth]{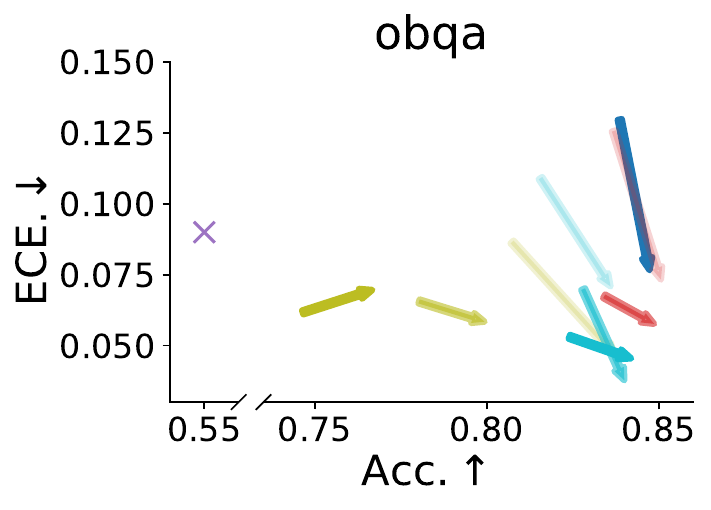}
    \end{subfigure}
    \begin{subfigure}[b]{0.325\linewidth}
   \includegraphics[width=\linewidth]{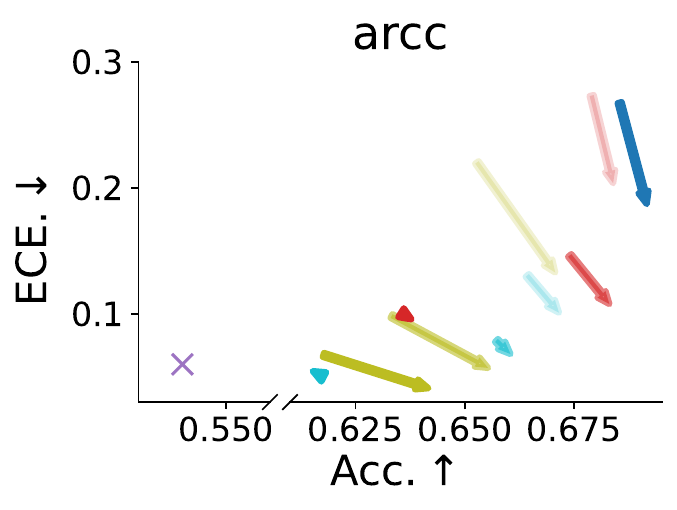}
    \end{subfigure}
    \begin{subfigure}[b]{0.325\linewidth}
              \includegraphics[width=\linewidth]{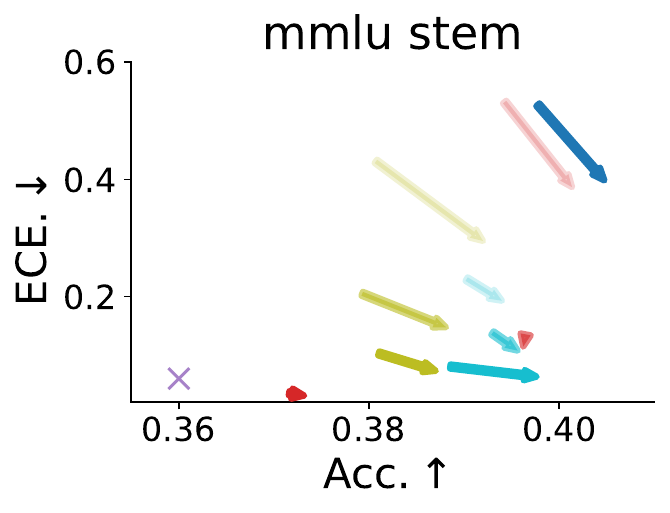}
    \end{subfigure}
    \vspace{-2pt}
    \caption{
    \textbf{\EOL improve both accuracy and calibration under different regularization techniques.}  Arrows link the performance of a single LoRA model (arrow tail) to the corresponding ensemble with 5 LoRA components (arrowhead), where the x-axis denotes validation accuracy and the y-axis expected calibration error. Arrow colors indicate regularization methods 
    %maja: this should be mentioned in the implementation details: (dark blue means the standard configuration of AdamW with no extra regularization), 
    and opacity reflects regularization strength. The majority of arrows are pointing toward the right bottom corner, suggesting that ensembling benefits both accuracy and calibration error measured by ECE.
    % Our proposed method: \EORL consistently reaches the right bottom corner, providing accurate and calibrated predictions.
    % \xw{TODO: }
    % Accuracy v.s. expected calibration error for different methods. Different points represent the value of metric averaged at different epochs and random seeds. Ensembled fine-tuning consistently improve accuracy for all methods. Different opacity represents different level of regularization: The darker the larger.
    % Standard (fine-tuning) shows high accuracy but bad calibration. Early stopping and weight decay has good calibration but decreased accuracy. By ensembling 
    }
    \label{fig:arrow_plots}
\end{figure}

 \vspace{-3pt}
\section{Related work}
 \vspace{-3pt}
% \paragraph{Ensembling of neural networks.} Deep ensemble is a commonly adopted technique for enhancing the robustness, generalization, and reliability of deep learning models~\citep{ovadia2019can}. The most standard way of constructing a deep ensemble would be to train the same model multiple times using different initialization~\citep{lakshminarayanan2017simple} or different hyper-parameter setting~\citep{wenzel2020hyperparameter}. It is also possible to use samples from the optimization trajectory~\citep{huang2017snapshot} or Bayesian posterior~\citep{neal2012bayesian,wenzel2020good,izmailov2021bayesian} to construct ensembles. However, it is not obvious to apply any of them in the LLM fine-tuning setting, as they all require storing and loading the full model checkpoints for predictions, which is not applicable LLMs.
% BatchEnsemble~\citep{wen2020batchensemble} performs deep ensemble by jointly training the origin model weights (slow matrix) and several rank-1 multiplicative matrices (fast matrix) that correspond to ensemble components, which controls the storage cost of ensembling. A follow-up work~\citep{tran2022plex} shows that it is possible to apply BatchEnsemble when training large-scale foundation models. However, it is unclear whether BatchEnsemble can be applied in LLM fine-tuning setting as it is unknown whether multiplicative low-rank matrices can be used as a method for fine-tuning out-of-box pre-trained LLM.

\textbf{Robust fine-tuning of language models.} A body of work has proposed regularization methods to improve generalization and calibration during fine-tuning of language models. 
For instance, \cite{he2022preserving} explores a mix of KL and L2 regularization on the extracted features to retain the calibration of pre-trained masked language models (MLM).
\cite{park2022calibration} introduces mixup~\citep{zhang2017mixup} into MLM fine-tuning, showcasing enhanced calibration at test time.
Our approach complements these methods: we can ensemble fine-tuning with any of these techniques (if compatible with LoRA adapters), 
and we will discuss such strategies extensively in later sections.

\textbf{Ensembling of Neural Networks.} Deep ensembles enhance the robustness and reliability of deep learning models~\citep{ovadia2019can}. 
Typically, they are constructed by training the same model with varied initializations~\citep{lee2015m,lakshminarayanan2017simple} or hyper-parameter settings~\citep{wenzel2020hyperparameter, zaidi2020neural}. 
Some methods use checkpoints from along the optimization trajectory~\citep{huang2017snapshot} or the Bayesian posterior~\citep{neal2012bayesian,Zhang2020Cyclical,wenzel2020good,izmailov2021bayesian}.
However, naively applying these methods in the LLM setting requires us to store and load complete model checkpoints which is impractical in many settings due to the very large memory requirements for storing multiple copies of an LLM.

\textbf{Ensembling in LLMs.} Two recent papers study ensembling for LLM fine-tuning \citep{gleave2022uncertainty,sun2022quantifying}.
However, these papers only consider full fine-tuning, optimizing all the weights, which requires them to store $M$ copies of the model, where $M$ is the number of ensemble components.
This is impractical for modern LLMs, so instead they are forced to work with smaller models; in particular, they work with GPT-2~\citep{radford2019language} with only 1.5 billion parameters. \cite{hewitt2021ensembles} consider an ensemble of LLMs consisting of two components: a model trained with full fine-tuning, and a model trained with LoRA. In contrast, we consider ensembling with a large number of LoRA components (e.g. 20) to improve accuracy and calibration.
% In contrast, our memory-efficient approach, \EOL, scales to larger LLMs (\LAMA), and allows for many ensemble components (Fig.~\ref{fig:extra_seed}).
% Another efficient LLM ensemble method is BatchEnsemble~\citep{wen2020batchensemble}. However, its efficiency comes from multiplicative adapters that in their current implementation can be used only during pre-training. It is unclear how to initialize BatchEnsemble with a pre-trained T5 model \citep{tran2022plex} without the adapters neutralizing the pre-trained weights in the fine-tuning tasks we consider in this paper. 
There also exists an efficient ensemble method BatchEnsemble~\citep{wen2020batchensemble}, where the ensemble components share a base model that is modified multiplicative by component-specific parameter-efficient rank-1 matrices, leading to reasonable storage demands comparable to our proposed \EOL. It has been applied on LLMs by \citet{tran2022plex} but for pre-training rather than fine-tuning.
% It concurrently trains original model weights with several randomly initialized rank-1 multiplicative matrices that correspond to ensemble components, which mitigates the storage demands. 
% However, it is unclear how to initialize and train BatchEnsemble without neutralizing the pre-trained weights in the fine-tuning settings we consider in this paper. 
% While it may be possible to adapt BatchEnsemble to the fine-tuning setting, this has not, to our knowledge, yet been considered, as it is unclear how to initialize and train BatchEnsemble without neutralizing the pre-trained weights in the fine-tuning settings. 
While it may be possible to adapt BatchEnsemble to the fine-tuning setting, this has not, to our knowledge, yet been considered. Indeed, we believe that such an adaptation is a non-trivial exercise that may require careful consideration of e.g.\ how to initialize and train the multiplicative adapters to avoid ``overpowering'' the pre-trained weights.

\textbf{Calibration and uncertainty quantification of LLMs.} Pre-trained LLMs already show reasonably good calibration \citep{OpenAI2023GPT4TR}.  
Nonetheless, there are several recent papers that seek to further enhance \emph{pre-trained} LLMs' calibration and uncertainty quantification ability in open-ended generation tasks.  In particular, \cite{lin2022teaching, kuhn2022clam} propose to use prompts to guide models to provide linguistically calibrated answers. 
\cite{zhou2023navigating} studies how LLMs express uncertainty in the natural language form. 
Our work is very different in that it focuses on mitigating very poor calibration that can emerge from fine-tuning.
% \cite{Li2020Rethinking, kumar2022fine-tuning, wortsman2021robust, he2023preserving, tian2023just} 

% \cite{hewitt2021ensembles,wortsman2021robust} consider ensemble of pre-trained model and fine-tuned model on the weight space, not ensemble of multiple models.

\section{Background}

\subsection{Fine-tuning large language models}
Fine-tuning assumes access to a pre-trained LLM, denoted by by $\W^*$, usually an auto-regressive model based on the transformer architecture. 
% We denote the parameters of the pre-trained model by $\W^*$. 
In the tasks we consider, the fine-tuning data consists of prompts, $\X = \{\x_n\}_{n=1}^N$, and answers $\y = \{y_n\}_{n=1}^N$, where the prompt can, e.\,g., describe a multiple-choice QA  problem (Fig.~\ref{fig:prompt_template}), and the answer can be in the label set $\mathcal{T} = \{``a", ``b", ``c", ``d"\}$, which is a subset of all tokens $\mathcal{V}$ the LLM can generate. 
Given this data, fine-tuning entails initializing the parameters at $\W = \W^*$ and minimizing the loss  $-\log p(\y \mid \X; \W)$.  In this paper, we consider tasks, where the label set $\mathcal{T}\subset\mathcal{V}$ of possible answers consists of single tokens.\footnote{Our method, LoRA ensembles also applies to open-ended generation tasks.} Typically, there will be tokens $v\notin\mathcal{T}$ with nonzero probability under the LLM. To study calibration accuracy and predictive uncertainty of LLM fine-tuning, we introduce the normalized task distribution
\begin{equation}
\label{eqn:task_norm}
p_\mathcal{T}(y \mid x_n; \W) = \biggl \{ \begin{aligned}
p(y \mid & x_n; \W)/Z_{\W} &&\text{if } \y \in \mathcal{T} \\
&0 &&\text{otherwise,}
\end{aligned}  \quad \text{ where } \quad Z_{\W}=\sum_{\y \in \mathcal{T}} p(y \mid x_n; \W).
\end{equation}
The normalized task distribution allows us to study the quality of predictions beyond accuracy.

\subsection{Deep ensembles}
A popular tool for improving the predictive uncertainty of deep learning methods are ensembles. Deep ensembles \citep{lakshminarayanan2017simple} offer a practical alternative for the fully Bayesian treatment of Bayesian neural networks \citep{neal2012bayesian} or Monte-Carlo Dropout \citep{gal2016dropout}. They simply average the predictions of $M$ networks which have been trained separately using different random initializations,
%Analogously, we can define an ensemble over LLMs, by averaging over $M$ components,  
\begin{equation}\label{eq:ens_formula}
    p_\text{ens}(y \mid \x_n) = \frac{1}{M} \sum_{m=1}^M p(y \mid \x_n; \W_m).
\end{equation}
%each of which is a LLM with separate parameters $\W_m$. Using an ensemble of LLMs, especially in the context of fine-tuning, presents the following two challenges:
%
%To address these challenges, we propose LoRA Ensembles. 

\subsection{Efficient Fine Tuning low-rank adapters (LoRA)}

LoRA \citep{hu2021lora} is a parameter-efficient fine-tuning technique. Instead of fine-tuning all the model weights, it learns additive correction terms, called {\em adapters}, whose low-rank structure greatly reduces the number of trainable parameters. 
%This makes LoRA a practical choice for LLM ensembles. 
%Each ensemble component will use the same pre-trained weights $\W^*$, but will have its own adapter term $\DW_m$.
%This introduces strong parameter sharing between the component-specific weights $\W_m = \W^* + \DW_m$, and adapter initialization as a source of randomness.
Each adapter $\DW = \alpha BA$ consists of trainable \emph{low-rank} matrices $ B\in \R^{d \times r}, A \in \R^{r \times k}$ of rank $r$ and a constant scaling factor $\alpha \in \R^+$, which is usually fixed. 
%In particular, following \citet{hu2021lora}, we apply the adapter only on the query and value matrix of the self-attention modules of the LLM (i.e. $W_q$ and $W_v$).
% [TODO: mention transformer architecture and which of the matrices actually have adapters]
%for each input feature vector $x \in \mathbb{R}^{d}$, we perform forward passing as
%\begin{equation}
%    h = (W + \DW) x = Wx + \DW x.
%\end{equation}
During fine-tuning, we fix $\W^*$ , and only optimize $\DW$:
\begin{equation}
\label{eq:objective}
  \mathcal{L}(\DW) = \min_{\DW} \sum_{n=1}^N -\log p(y_n \mid \x_n ; \W^*+\DW).
\end{equation}
Critically, $\DW$ represents far fewer parameters, so is much easier to fine-tune in constrained compute environments. As suggested by ~\citep{hu2021lora}, it is common to initialize $A$ randomly and $B$ as zero, in that way, we can have $\W^* + \DW = \W^*$ at the beginning of the optimization, i.e. the fine-tuned model starts from the pre-trained model.

\subsection{Regularization}

\textbf{Output space regularization via KL regularization.}  In LLM fine-tuning, it is common to include a KL regularization~\citep{schulman2017proximal,bai2022training,ouyang2022training,korbak2022rl, he2022preserving} to make the output distribution of the fine-tuned model close to that of the pre-trained model. In our setting, we consider the following KL regularization objective
\begin{equation}\label{eq:kl_objective}
    \beta D_\textit{KL}\infdivx{p(\y \mid \X; W, \DW)}{p(\y \mid \X; W)},
\end{equation}
which is added to Eq.~\eqref{eq:objective} during optimization where $\beta$ controls the strength of the regularization.
% for all experiments, we consider $\beta$ in the range $\{1.0, 0.5, 0.1, 0.025, 0.01\}$

\textbf{Implicit regularization via early stopping.} Another commonly adopted regularization method is early stopping. Early stopping halts the optimization when certain criteria are met such that the model is not over-optimized. The fewer epochs used, the stronger the regularization is. 
% In particular, we consider early stopping after certain epochs ranging from $1$ to $3$: 
% \xw{Remove the hyper parame}
\section{Method}
We propose \EOL, an ensemble of LLMs where each of the ensemble components is fine-tuned with LoRA. Remember that LoRA learns only low-rank additive correction terms, called {\em adapters}, with a very small number of parameters.
This makes LoRA a practical choice for LLM ensembles. 
Each ensemble component will use the same pre-trained weights $\W^*$, but will have its own adapter term $\DW_m$.
This introduces strong parameter sharing between the component-specific weights $\W_m = \W^* + \DW_m$, and the adapter initialization becomes a source of diversity.
After fine-tuning, LoRA facilitates efficient storage and retrieval, empowering us to swiftly execute ensembles at the prediction stage. Crucially, at test time, the large base model $\W^*$ is loaded only once, while the low-rank adapters can be loaded and unloaded with negligible overhead.

Importantly, \EOL can be applied on top of modified fine-tuning protocols, that include regularization. 
For example, these regularization methods might keep all the fine-tuned models in the ensemble close to the pre-trained model, which may additionally improve calibration over just the improvement from ensembling.
In extensive experiments (Sec.~\ref{sec:exp_results}), we find that ensembling can offer additional improvements in accuracy and calibration over those offered by regularization alone.
Finally, we consider an additional form of regularization that we were not able to find explicitly discussed in prior work.
%The ensembled predictive distribution (Eq.~\ref{eq:ens_formula}) usually shows better calibration than a single base model in that the sharpness of each overconfidence component gets averaged out through ensembling. However, we notice that the average-out effect of ensembling is not sufficient for solving the overconfidence. Therefore, we consider combining regularization techniques with \EOL, which we notice plays a critical role in improving the calibration of ensembling, especially when the training data is small or on out-of-distribution samples. Based on the observation that, the pre-trained model usually make better-than-random-guess and calibrated prediction (Table.~\ref{table:few_shot}), we consider regularization techniques that enforce the fine-tuned model to stay close to the pre-trained model.
%\subsection{Regularized Ensemble Components}
%\label{sec:reg}
%We study the effect of various regularization techniques on LoRA ensembles, including weight space regularization, output space regularization and implicit regularization.
%
%\paragraph{
In particular, we consider penalizing the LoRA $B$ matrix by including a very large %\textcolor{olive}{
 weight decay
%} 
 term in AdamW \citep{loshchilov2018decoupled}. At the $t_\textit{th}$ time step, the AdamW update for $B$ is
\begin{equation}\label{eq:weight_decay}
    B_t \leftarrow B_{t - 1} - \gamma (g_{t - 1} - \lambda B_{t-1}).
\end{equation}
Here, $\gamma$ is the step size, $g_{t-1}$ is the normalized gradient acquired from standard Adam~\citep{kingma2014adam} with no regularization and $\lambda$ adjusts the strength of regularization. Although weight decay is a very standard regularization technique, we find that the usual setup with a $\lambda$ of $1\mathrm{e}{-2}$ (the default setting from PyTorch's AdamW, denoted as ``None'' in our paper) barely helps resolve the overconfidence issue. Instead, we adopt an extreme large value of $\lambda$ ranging from $1\mathrm{e}{2}$ to  $1\mathrm{e}{3}$. In addition, we adopt weight decay only on the $B$ matrix of $\DW$, which we notice shows the best performance (Appendix.~\ref{appendix:weight_decay}).

\section{Empirical Study of LoRA Ensembles}\label{sec:experiment}

In this section, we evaluate \EOL on a collection of datasets to show its benefits in various QA settings (described in Sec.~\ref{sec:setup}), and find that it leads to better predictive and calibration accuracy than baseline approaches (Sec.~\ref{sec:exp_results}).
In addition, we study the effect of regularization in \EOL, where we find that \EOL is complementary to many regularization techniques.
% at the cost of sacrificing a little accuracy.
Finally, we conduct ablation studies, to better understand the effect of our modeling choices on ensemble diversity and \EOL's performance (Sec.~\ref{sec:ablation}).  

% \parhead{Goals of the study}
% \begin{itemize}[noitemsep,topsep=0pt,parsep=0pt,partopsep=0pt]
% \item We compare \EOL with two baseline ensemble methods (\cref{sec:other_ens_method}): Last-layer ensembles and Monte Carlo dropout where we observe \EOL show the best performance under the same number of forward passes.
% \item We compare \EORL with the un-regularized version (\cref{sec:reg}) where we show that regularization significantly improves the ECE. and NLL while losing some accuracy.
% \item We vary the number of ensemble components (\cref{sec:ens_number_ablation}) where we find increasing the number of components increases the performance of \EOL but still performs worse than the regularized version. 
% \item We perform \EOL with controlled the source of randomness (\cref{sec:random_source_ablation}), in which we find both random initialization and data shuffling contribute to the diversity in ensembling.
% \item We vary the level of regularization strength (\cref{sec:reg_strength_ablation}) and we notice a high degree of regularization could potentially cause \EOL to become less diverse and thus less effective.
% \end{itemize}

\subsection{Experimental Set-up}
\label{sec:setup}
\textbf{Multiple-Choice Question Answering Datasets.} For our experiments, we choose six popular multiple-choice QA  datasets for evaluation: CommonsenseQA~\citep[cqa,][]{talmor-etal-2019-commonsenseqa}, OpenBook~\citep[obqa,][]{OpenBookQA2018}, social sciences (mmlu ss.) and STEM (mmlu stem) subset from MMLU~\citep{hendryckstest2021}, ARC-easy (arce) and ARC-challenge (arcc) from AI2 Reasoning Challenge~\citep{allenai:arc}. Questions in cqa have 5 options while the others all have 4 options. We provide details for the training and validation set of each task in Table~\ref{table:dataset_info} in the appendix, we provide example questions for each task in Appendix~\ref{sec:question_sample}. 

\textbf{Evaluation metrics.} For all 6 tasks, we first measure the accuracy (Acc.) on the validation set. However, problems such as bad calibration or lack of uncertainty quantification can not be reflected through accuracy. Therefore we incorporate negative log-likelihood (NLL.), which measures the model uncertainty on held-out validation datasets, and expected calibration error~\citep[ECE.,][]{guo2017calibration} which assesses the alignment between predicted probabilities and actual empirical accuracy. Since safe deployment in real-world applications requires models to behave predictably when the data comes from another domain, we also study OOD performance.  In particular, we test models fine-tuned on cqa on test samples from mmlu as OOD and we test models fine-tuned on a subset of mmlu with test samples from other mmlu subcategories. We then compute the accuracy, NLL., ECE., and additionally, the OOD detection performance measured by AUROC on the OOD test samples using negative maximum softmax probability~\citep{hendrycks2016baseline} as the score.

\textbf{Implementation Details of LoRA Ensembles.} We build LoRA ensembles by fine-tuning \LAMA~\citep{touvron2023llama} which has 13 billion parameters, where we use Transformers~\citep{wolf-etal-2020-transformers} and Huggingface PEFT~\citep{peft} for model and LoRA implementations. In most experiments, we build ensembles with $M=5$ components, though our ablation study also contains larger ensembles. As in \citet{hu2021lora}, we apply the adapter $\DW = \alpha BA$ only on the query and value matrices of the self-attention modules of \LAMA and we fix $\alpha=32$. With rank $r = 8$, each ensemble component has 6 million trainable parameters. The adapter matrices $B$ are initialized to be zero, while the entries of $A$ are randomly initialized using Kaiming Uniform~\citep{he2015delving}. We use AdamW for all experiments and run optimizations for 20 epochs with a fixed step size of $5\mathrm{e}{-5}$ . We use a batch size of 32 for cqa, 16 for obqa, arcc, and arce, and 8 for mmlu ss. and stem. Half-precision is used for all the forward and backward passes after which we convert the output logits to single precision when computing metrics. For all datasets, we experiment with four variations of \EOL: Default AdamW configuration with $\gamma=0.01$, denoted as \textcolor{blue}{\textbf{None}} in the figures; \textcolor{red}{\textbf{KL}} regularization from Eq.~\eqref{eq:kl_objective} with $\beta \in \{0.01, 0.05, 0.1\}$; \textcolor{cyan}{\textbf{Early stopping}} after $\{1,2,3\}$ epochs, and very large \textcolor{olive}{\textbf{weight decay}} on $B$ from Eq.~\eqref{eq:weight_decay} with $\gamma \in \{1\mathrm{e}{2}, 5\mathrm{e}{2}, 1\mathrm{e}{3}\}$.

We consider the following approaches as baselines

\noindent\begin{itemize}[leftmargin=*, topsep=0pt,parsep=0pt]
\item \textbf{LoRA (M=1)} For all variations of \EOL, we report the averaged performance of the \emph{single} ensemble members. We represent the results with solid lines in trace figures, in contrary to dashed lines for the ensembled versions.
\item \textbf{Few shot} For each question in the validation set, we append $(\X,\y)$ pairs from the training set in front of the prompts as ``demonstration'' to perform few shot learning~\citep{brown2020language}. In our experiments, we randomly draw $3$ pairs of $(\X,\y)$ without replacement and evaluate the performance through the average of 10 random draws. We perform few shot experiments only on the pre-trained model without any fine-tuning.
\item \textbf{Last-layer ensembles} Last-layer fine-tuning, also known as linear probing~\citep{du2021fewshot,Wu2020Understanding}, refers to freezing the model but only fine-tuning the last linear layer. We fine-tuned the rows in the linear head that correspond to the token for the options. multiple times starting from the pre-trained weights under different random seeds to construct an ensemble.
\item \textbf{Monte Carlo (MC) dropout.} When dropout is employed at training time, we can use MC dropout~\citep{gal2016dropout} to perform ensembling: Instead of training multiple LoRA adapters, we can train a single one and keep Dropout on at test time and perform multiple forward passes with nodes randomly shut down. MC dropout has previously been adopted in masked language models for incorporating uncertainty estimation~\citep{sankararaman2022bayesformer,vazhentsev2022uncertainty}. We combine dropout with \emph{standard} LoRA fine-tuning by adding dropout on the input of the LoRA adapter following the implementation of \cite{peft}. 
\end{itemize}

% - Last-layer ensemble 
% Last-layer fine-tuning refers to fine-tuning only the rows in the final linear head that correspond to the token for the options. We fine-tuned the linear head multiple times starting from the pre-trained weights under different random seeds to construct an ensemble. 

% For all ensemble methods, we make predictions with 5 ensemble components.

\subsection{Results}\label{sec:exp_results}

In Fig.~\ref{fig:arrow_plots}, we present the validation accuracy and ECE. of different \EOL fine-tuning methods after 20 epochs.
Critically, ensembling usually gives considerable improvements in accuracy and calibration compared with the single-component versions regardless of the regularization method or strength. 
\EOL also shows significantly improved accuracy compared with few shot learning (purple ``x''), confirming the value of fine-tuning.
Fig.~\ref{fig:arrow_plots} also shows that regularization usually improves calibration, as measured by ECE.
However, the effect of regularization on accuracy is more inconsistent: stronger regularization often reduces accuracy (e.g.\ stronger regularization in arce reduces accuracy) though sometimes increases accuracy (mmlu stem; early stopping).

\begin{figure}[t]
    \centering
    \begin{subfigure}[b]{0.8\linewidth}
        \centering
        \includegraphics[width=\linewidth]{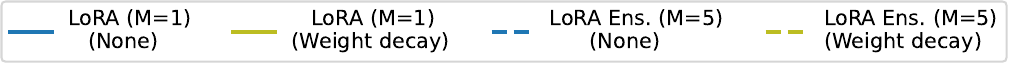}
        \caption*{} % The '*' ensures there's no figure numbering for the legend
    \end{subfigure} \\[-3.8ex]
    \centering
    \begin{subfigure}[b]{\linewidth}
        \includegraphics[width=\linewidth]{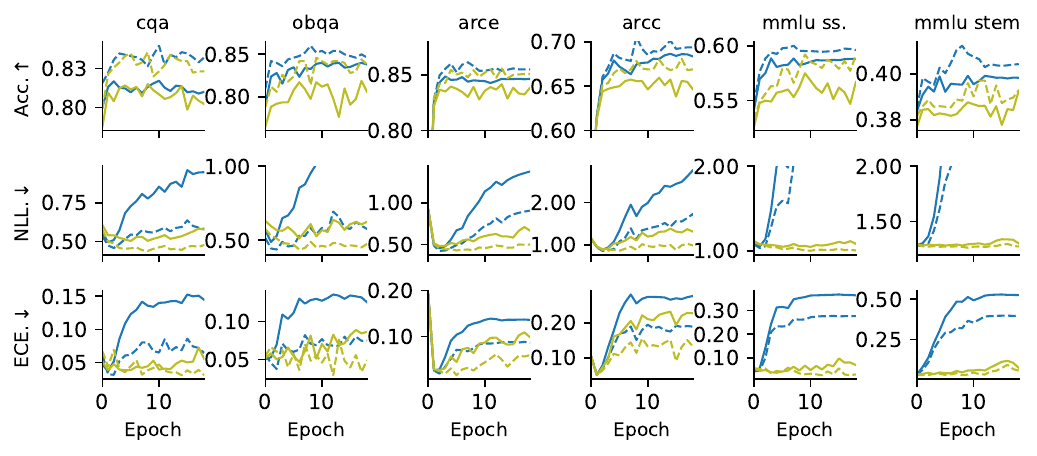}
    \end{subfigure}
    \caption{\textbf{\EOL improves accuracy while regularization prevents  NLL from blowing up.} For all ensemble results we use $M=5$ components. We use $\lambda=1\mathrm{e}{3}$ for mmlu subsets and $\lambda=1\mathrm{e}{2}$ for others for weight decay.}
    \label{fig:trace_standard_wd}
\end{figure}

Next, we look at the behavior of (regularized) \EOL across training (Fig.~\ref{fig:trace_standard_wd}).
This reinforces the results from Fig.~\ref{fig:arrow_plots}.
In particular, \EOL consistently improves both calibration and accuracy, whether applied with or without regularization (here, weight decay).
However, weight decay has conflicting effects: it seems to usually reduce accuracy while improving calibration.
Interestingly, the NLL metric seems to become very large for several of the datasets (e.g.\ mmlu ss. and stem).
This is likely because the NLL heavily penalizes overconfidence: assigning very, very low probability to the right answer.
Interestingly, ensembling on its own was not sufficient to prevent this dramatic increase in NLL, while weight decay was sufficient to prevent the dramatic increase (though weight decay in combination with ensembling consistently gave the best NLL).

\begin{figure}
    \centering
    \begin{subfigure}[b]{0.5\linewidth}
        \centering
        \includegraphics[width=\linewidth]{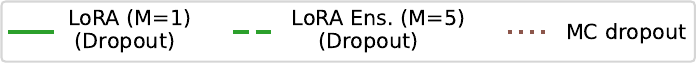}
        \caption*{} % The '*' ensures there's no figure numbering for the legend
    \end{subfigure}\\[-3.8ex] 
    \centering
    \begin{subfigure}[b]{\linewidth}
    \includegraphics[width=\linewidth]{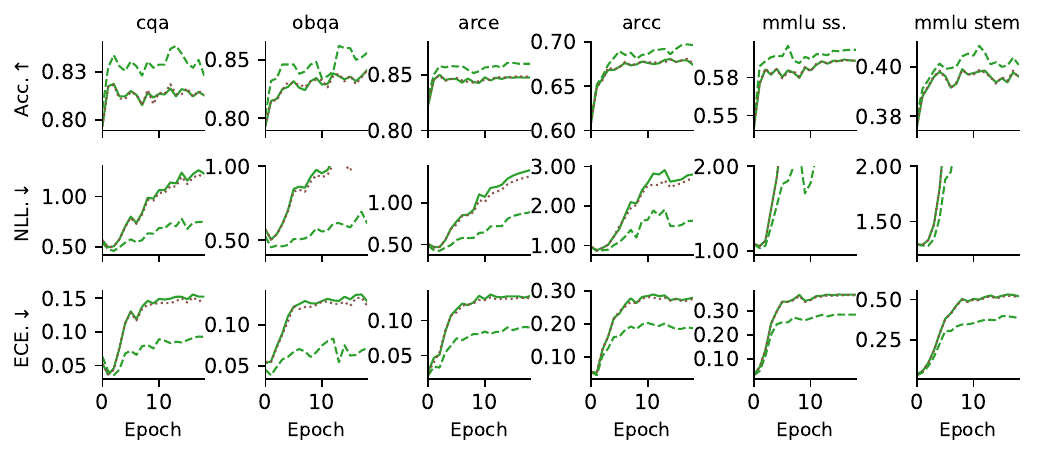}
    \end{subfigure}
     \caption{
        \textbf{Ensemble of LoRA significantly outperforms MC dropout under the same number of ensemble members.} When employing dropout during the fine-tuning, an alternative ensemble strategy becomes available: Keeping dropout on at test time to implement Monte Carlo (MC) dropout. However, MC dropout offers only marginal performance gains compared to a standalone model, outperformed by ensembles of independently trained LoRA models when both methods employ the same number of ensemble members (chosen as 5 in our experiments).}
    \label{fig:dropout}
\end{figure}

\begin{figure}[t]
    \centering
    \begin{subfigure}[b]{0.8\linewidth}
        \centering
     \includegraphics[width=\linewidth]{figs/big_row_legend_new_v2.pdf}    
        \caption*{} 
    \end{subfigure}\\[-3ex] 
    \centering
     \begin{subfigure}[b]{\linewidth}
        \includegraphics[width=\linewidth]{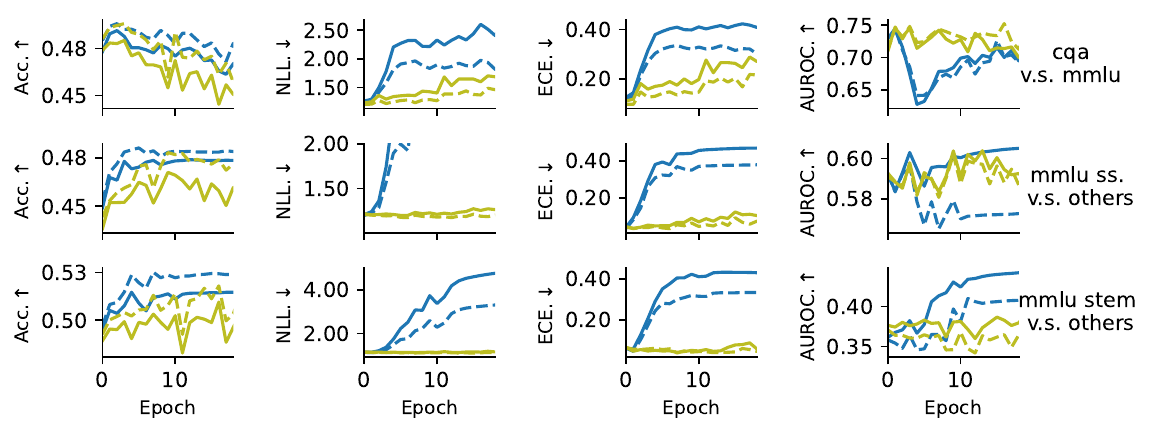}
    \end{subfigure}
    \caption{\textbf{Ensembles offer benefits for accuracy and calibration over regularized and unregularized fine-tuning approaches in OOD settings.}
    % On the first row, all models show degraded accuracy on mmlu as fine-tuning continues however the un-regularized version shows increased NLL. and ECE. 
    Note that all methods show AUROC around or lower than $0.5$ on the second and third row, we suspect the models would fail to \emph{detect} OOD samples if they can \emph{generalize} to them, as the accuracy increases throughout fine-tuning.
    }
    \label{fig:ood}
\end{figure}

% \textbf{MC dropout.} \citep{gal2016dropout} 
% Next, we consider MC dropout, an alternative way of creating an ``ensemble''.  
% Specifically, in MC dropout, we keep dropout on at test-time, then run the model multiple times, and average the output probabilities.  
We present the performance of MC dropout in Fig.~\ref{fig:dropout}. We find that MC dropout shows a marginal improvement over the performance of a single model while \EOL gives dramatically larger improvements in terms of both accuracy and calibration. The performance of last-layer ensembles is presented in Table~\ref{table:few_shot_and_lle}, which also shows worse accuracy than \EOL. Fine-tuning only the linear head might not be expressive enough for downstream tasks adaptation.
%
% \textbf{Last-layer ensembles.} \citep{du2021fewshot, Wu2020Understanding} An additional alternative is to ...
%
Lastly, we find that ensembling is also helpful in the OOD setting (Fig.~\ref{fig:ood}). While it fails to resolve catastrophic forgetting in cqa v.s. mmlu, it shows improvements both in terms of NLL. and ECE., providing more reliable predictions on unseen domains.

%\textbf{Effect of regularization on \EOL.} 
%% Regularization is an important ingredient of \EOL.
%
%In Fig.~\ref{fig:trace_standard_wd}, we notice that the NLL of LoRA Ens. (M=5) (None) blows up on small datasets, while using weight decay regularization alleviates the problem. 
%In Fig.~\ref{fig:ood}, we notice that regularization can also help better detect OOD samples in cases when the model fails to generalize.
%However, we do notice that regularization could affect the diversity of the ensemble, something critical for the ensemble's performance~\citep{breiman2001random}. In Fig.~\ref{fig:arrow_plots}, high strength of KL regularization and early stopping could cause the performance gain of ensembling to vanish. This is not surprising in that KL regularization directly forces all ensemble components to make predictions similar to the pre-trained model while early stopping prevents different ensemble models from further moving away from the pre-trained model. Weight decay suffers the least from this problem, we suspect that this is caused by the complicated relationship between the weight space and the output space as well as we are preforming optimization long enough for ensemble members to diverge.

\begin{table}
\centering
\footnotesize
\setlength\tabcolsep{3pt}
\caption{Performance of last-layer Ensembles. Last-layer Ensembles underfit the data, showing accuracy significantly worse than \EOL.}
\begin{tabular}{lcccccc}
\toprule
 & \multicolumn{2}{c}{Acc. $\uparrow$} & \multicolumn{2}{c}{NLL. $\downarrow$} & \multicolumn{2}{c}{ECE. $\downarrow$} \\
 & LoRA Ens. & Last-layer Ens. & LoRA Ens. & Last-layer Ens. & LoRA Ens. & Last-layer Ens. \\
\hline
cqa & 0.83 & 0.52 & 0.58 & 1.25 & 0.06 & 0.06 \\
obqa & 0.85 & 0.48 & 0.58 & 1.27 & 0.07 & 0.12 \\
arce & 0.86 & 0.72 & 0.92 & 0.79 & 0.09 & 0.06 \\
arcc & 0.69 & 0.48 & 1.46 & 1.30 & 0.19 & 0.15 \\
mmlu ss. & 0.60 & 0.46 & 2.72 & 1.47 & 0.28 & 0.19 \\
mmlu stem & 0.41 & 0.32 & 4.04 & 1.72 & 0.40 & 0.25 \\
\bottomrule
\end{tabular}
\label{table:few_shot_and_lle}
\end{table}

\subsection{Additional Ablations}\label{sec:ablation}

\begin{figure}[t]
    \centering
    \begin{subfigure}[b]{0.5\linewidth}
        \centering
        \includegraphics[width=\linewidth]{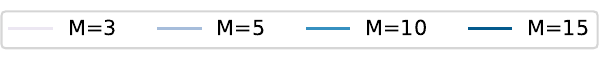}
        \caption*{} % The '*' ensures there's no figure numbering for the legend
    \end{subfigure}\\[-3ex] 
    \centering
    \begin{subfigure}[b]{\linewidth}
        \includegraphics[width=\linewidth]{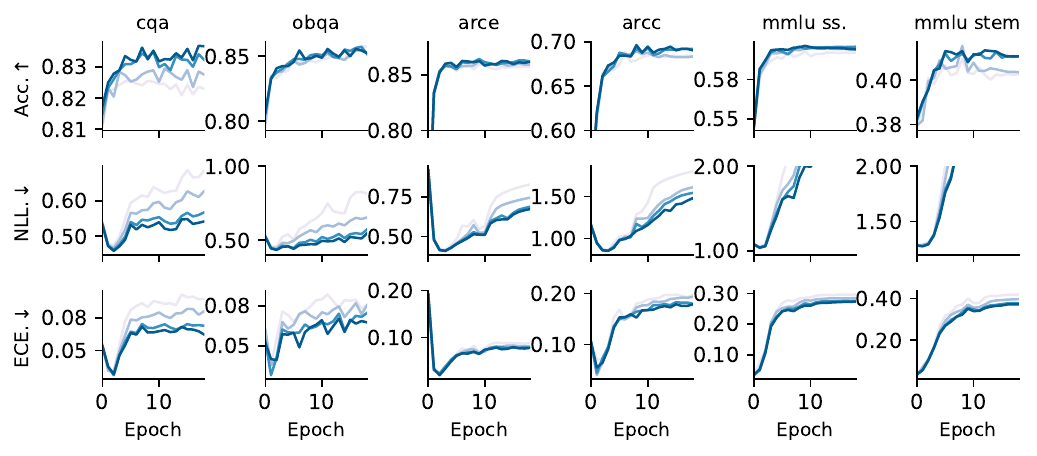}
    \end{subfigure}
    \caption{\textbf{Increasing the number of ensembles improves accuracy and calibration.} With \EOL, we can efficiently ensemble with a large number of components, however, the performance gains from increasing the number of components become less substantial with larger $M$.}
    \label{fig:extra_seed}
\end{figure}

\begin{figure}[t]
    \centering
    \begin{subfigure}[b]{0.8\linewidth}
        \centering
        \includegraphics[width=\linewidth]{figs/big_row_legend_new_v2.pdf}
        \caption*{} % The '*' ensures there's no figure numbering for the legend
    \end{subfigure}\\[-3ex] 
    \centering
        \begin{subfigure}[b]{0.325\linewidth}
        \includegraphics[width=\linewidth]{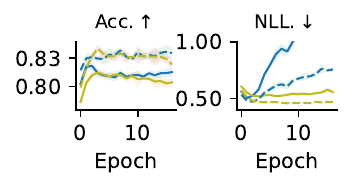}
        \caption{Dataset shuffling only.}
    \end{subfigure}
    \centering
    \begin{subfigure}[b]{0.325\linewidth}
        \includegraphics[width=\linewidth]{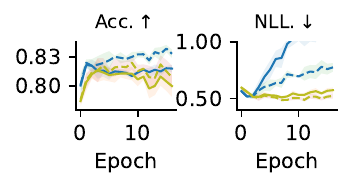}
        \caption{Random initialization only}
    \end{subfigure}
    \begin{subfigure}[b]{0.325\linewidth}
        \includegraphics[width=\linewidth]{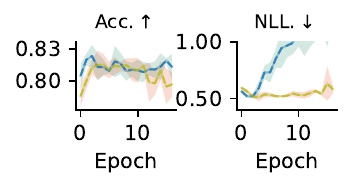}
        \caption{No randomness}
    \end{subfigure}
    \caption{\textbf{Randomness from initialization and dataset shuffling both contribute to the diversity of LoRA ensembles.} The diversity of ensembles can be reflected by the gap between the dashed lines and solid lines. When regularization is used, SGD noise from dataset shuffling alone is more beneficial than random initialization alone.}
    \label{fig:control_randomness}
\end{figure}

% In this section, we provide a more detailed understanding of LoRA ensembles. 
In this section, we perform extra ablations to gain a better understanding of the effect of the number of ensemble components and the randomness sources on the performance \EOL.

\textbf{Number of ensemble components.} 
To start with, we study the effects of ensemble component number $M$, we show the results in Fig.~\ref{fig:extra_seed} where we experiment with \EOL under different numbers of ensemble components. We collect 20 ensemble components in total and we report the average results of 5 random draws from them for each $M$.
% For all numbers, the results reported are based on 5 independent trials.
We notice that increasing $M$ improves all metrics however the marginal benefit of increasing components diminishes as $M$ becomes larger.
% However, using an extra number of ensemble components does not resolve the overconfidence problem, which again emphasizes the importance of introducing regularization into fine-tuning.

\parhead{Ensemble diversity under different regularization strengths.} In Fig.~\ref{fig:arrow_plots}, high strength of KL regularization and early stopping could cause the performance gain of ensembling to vanish. This is not surprising in that KL regularization directly forces all ensemble components to make predictions similar to the pre-trained model while early stopping prevents different ensemble models from further moving away from the pre-trained model. Weight decay suffers the least from this problem, we suspect that this is caused by the complicated relationship between the weight space and the output space as well as we are performing optimization long enough for ensemble members to diverge.

\parhead{Ensemble diversity under different sources of randomness.}
Next, we study the source of randomness in LoRA ensembles. As discussed in previous sections, the diversity of LoRA ensembles comes from two sources: The random initialization and the randomness from dataset shuffling (i.e. SGD noise).
It is often observed that random initialization contributes mostly to the diversity of ensemble~\citep{fort2019deep}. However, it is unclear whether this is the case for \EOL. To investigate, we conduct experiments on cqa under three settings: Dataset shuffling with fixed initialization, random initialization with fixed dataset shuffling, and no randomness (both fixed). For each setting, we conduct experiments with 5 independent trials and the results are presented in Fig.~\ref{fig:control_randomness}. We observe that LoRA ensembles can work with either source of randomness alone, while randomness from dataset shuffling (i.e. SGD noise) contributes more to the ensemble performance.
However, it is hard to decompose exactly the contribution of each source of randomness, and in practice, one should incorporate both random initialization and dataset shuffling for better diversity.

\section{Discussion}
In this paper, we develop a new method for LLM fine-tuning: \EOL. Our empirical results on 6 datasets demonstrate that our proposed method improves both the accuracy and calibration of fine-tuning a single model. In addition, we propose to combine regularization techniques together with ensembling for better calibration.
Broadly, LLMs have demonstrated their power in a variety of scenarios, but their safety issues have started to draw more and more attention~\citep{wei2023jailbroken,jones2022capturing,perez2022red}: Real-world applications require not only the LLM to be accurate but also reliable. Our method provides a key ingredient towards addressing these concerns, in that it helps LLMs to make not only precise but also calibrated predictions.

% enables safer deployment of fine-tuned LLMs in real-world applications. 

\newpage
\bibliographystyle{iclr2024_conference}
\bibliography{main}

\begin{thebibliography}{56}
\providecommand{\natexlab}[1]{#1}
\providecommand{\url}[1]{\texttt{#1}}
\expandafter\ifx\csname urlstyle\endcsname\relax
  \providecommand{\doi}[1]{doi: #1}\else
  \providecommand{\doi}{doi: \begingroup \urlstyle{rm}\Url}\fi

\bibitem[Bai et~al.(2022)Bai, Jones, Ndousse, Askell, Chen, DasSarma, Drain, Fort, Ganguli, Henighan, et~al.]{bai2022training}
Yuntao Bai, Andy Jones, Kamal Ndousse, Amanda Askell, Anna Chen, Nova DasSarma, Dawn Drain, Stanislav Fort, Deep Ganguli, Tom Henighan, et~al.
\newblock Training a helpful and harmless assistant with reinforcement learning from human feedback.
\newblock \emph{arXiv preprint arXiv:2204.05862}, 2022.

\bibitem[Breiman(2001)]{breiman2001random}
Leo Breiman.
\newblock Random forests.
\newblock \emph{Machine learning}, 45:\penalty0 5--32, 2001.

\bibitem[Brown et~al.(2020)Brown, Mann, Ryder, Subbiah, Kaplan, Dhariwal, Neelakantan, Shyam, Sastry, Askell, et~al.]{brown2020language}
Tom Brown, Benjamin Mann, Nick Ryder, Melanie Subbiah, Jared~D Kaplan, Prafulla Dhariwal, Arvind Neelakantan, Pranav Shyam, Girish Sastry, Amanda Askell, et~al.
\newblock Language models are few-shot learners.
\newblock \emph{Advances in neural information processing systems}, 33:\penalty0 1877--1901, 2020.

\bibitem[Chung et~al.(2022)Chung, Hou, Longpre, Zoph, Tay, Fedus, Li, Wang, Dehghani, Brahma, et~al.]{chung2022scaling}
Hyung~Won Chung, Le~Hou, Shayne Longpre, Barret Zoph, Yi~Tay, William Fedus, Eric Li, Xuezhi Wang, Mostafa Dehghani, Siddhartha Brahma, et~al.
\newblock Scaling instruction-finetuned language models.
\newblock \emph{arXiv preprint arXiv:2210.11416}, 2022.

\bibitem[Clark et~al.(2018)Clark, Cowhey, Etzioni, Khot, Sabharwal, Schoenick, and Tafjord]{allenai:arc}
Peter Clark, Isaac Cowhey, Oren Etzioni, Tushar Khot, Ashish Sabharwal, Carissa Schoenick, and Oyvind Tafjord.
\newblock Think you have solved question answering? try arc, the ai2 reasoning challenge.
\newblock \emph{arXiv:1803.05457v1}, 2018.

\bibitem[Du et~al.(2021)Du, Hu, Kakade, Lee, and Lei]{du2021fewshot}
Simon~Shaolei Du, Wei Hu, Sham~M. Kakade, Jason~D. Lee, and Qi~Lei.
\newblock Few-shot learning via learning the representation, provably.
\newblock In \emph{International Conference on Learning Representations}, 2021.
\newblock URL \url{https://openreview.net/forum?id=pW2Q2xLwIMD}.

\bibitem[Fort et~al.(2019)Fort, Hu, and Lakshminarayanan]{fort2019deep}
Stanislav Fort, Huiyi Hu, and Balaji Lakshminarayanan.
\newblock Deep ensembles: A loss landscape perspective.
\newblock \emph{arXiv preprint arXiv:1912.02757}, 2019.

\bibitem[Gal \& Ghahramani(2016)Gal and Ghahramani]{gal2016dropout}
Yarin Gal and Zoubin Ghahramani.
\newblock Dropout as a bayesian approximation: Representing model uncertainty in deep learning.
\newblock In \emph{international conference on machine learning}, pp.\  1050--1059. PMLR, 2016.

\bibitem[Gleave \& Irving(2022)Gleave and Irving]{gleave2022uncertainty}
Adam Gleave and Geoffrey Irving.
\newblock Uncertainty estimation for language reward models.
\newblock \emph{arXiv preprint arXiv:2203.07472}, 2022.

\bibitem[Guo et~al.(2017)Guo, Pleiss, Sun, and Weinberger]{guo2017calibration}
Chuan Guo, Geoff Pleiss, Yu~Sun, and Kilian~Q Weinberger.
\newblock On calibration of modern neural networks.
\newblock In \emph{International conference on machine learning}, pp.\  1321--1330. PMLR, 2017.

\bibitem[He et~al.(2022)He, Chen, and Zhu]{he2022preserving}
Guande He, Jianfei Chen, and Jun Zhu.
\newblock Preserving pre-trained features helps calibrate fine-tuned language models.
\newblock In \emph{The Eleventh International Conference on Learning Representations}, 2022.

\bibitem[He et~al.(2015)He, Zhang, Ren, and Sun]{he2015delving}
Kaiming He, Xiangyu Zhang, Shaoqing Ren, and Jian Sun.
\newblock Delving deep into rectifiers: Surpassing human-level performance on imagenet classification.
\newblock In \emph{Proceedings of the IEEE international conference on computer vision}, pp.\  1026--1034, 2015.

\bibitem[Hendrycks \& Gimpel(2016)Hendrycks and Gimpel]{hendrycks2016baseline}
Dan Hendrycks and Kevin Gimpel.
\newblock A baseline for detecting misclassified and out-of-distribution examples in neural networks.
\newblock \emph{arXiv preprint arXiv:1610.02136}, 2016.

\bibitem[Hendrycks et~al.(2021)Hendrycks, Burns, Basart, Zou, Mazeika, Song, and Steinhardt]{hendryckstest2021}
Dan Hendrycks, Collin Burns, Steven Basart, Andy Zou, Mantas Mazeika, Dawn Song, and Jacob Steinhardt.
\newblock Measuring massive multitask language understanding.
\newblock \emph{Proceedings of the International Conference on Learning Representations (ICLR)}, 2021.

\bibitem[Hewitt et~al.(2021)Hewitt, Li, Xie, Newman, and Liang]{hewitt2021ensembles}
John Hewitt, Xiang~Lisa Li, Sang~Michael Xie, Benjamin Newman, and Percy Liang.
\newblock Ensembles and cocktails: Robust finetuning for natural language generation.
\newblock 2021.

\bibitem[Hu et~al.(2021)Hu, Shen, Wallis, Allen-Zhu, Li, Wang, Wang, and Chen]{hu2021lora}
Edward~J Hu, Yelong Shen, Phillip Wallis, Zeyuan Allen-Zhu, Yuanzhi Li, Shean Wang, Lu~Wang, and Weizhu Chen.
\newblock Lora: Low-rank adaptation of large language models.
\newblock \emph{arXiv preprint arXiv:2106.09685}, 2021.

\bibitem[Huang et~al.(2017)Huang, Li, Pleiss, Liu, Hopcroft, and Weinberger]{huang2017snapshot}
Gao Huang, Yixuan Li, Geoff Pleiss, Zhuang Liu, John~E. Hopcroft, and Kilian~Q. Weinberger.
\newblock Snapshot ensembles: Train 1, get m for free.
\newblock In \emph{International Conference on Learning Representations}, 2017.
\newblock URL \url{https://openreview.net/forum?id=BJYwwY9ll}.

\bibitem[Izmailov et~al.(2021)Izmailov, Vikram, Hoffman, and Wilson]{izmailov2021bayesian}
Pavel Izmailov, Sharad Vikram, Matthew~D Hoffman, and Andrew Gordon~Gordon Wilson.
\newblock What are bayesian neural network posteriors really like?
\newblock In \emph{International conference on machine learning}, pp.\  4629--4640. PMLR, 2021.

\bibitem[Jones \& Steinhardt(2022)Jones and Steinhardt]{jones2022capturing}
Erik Jones and Jacob Steinhardt.
\newblock Capturing failures of large language models via human cognitive biases.
\newblock \emph{Advances in Neural Information Processing Systems}, 35:\penalty0 11785--11799, 2022.

\bibitem[Kingma \& Ba(2014)Kingma and Ba]{kingma2014adam}
Diederik~P Kingma and Jimmy Ba.
\newblock Adam: A method for stochastic optimization.
\newblock \emph{arXiv preprint arXiv:1412.6980}, 2014.

\bibitem[Kojima et~al.(2022)Kojima, Gu, Reid, Matsuo, and Iwasawa]{kojima2022large}
Takeshi Kojima, Shixiang~Shane Gu, Machel Reid, Yutaka Matsuo, and Yusuke Iwasawa.
\newblock Large language models are zero-shot reasoners.
\newblock \emph{Advances in neural information processing systems}, 35:\penalty0 22199--22213, 2022.

\bibitem[Korbak et~al.(2022)Korbak, Perez, and Buckley]{korbak2022rl}
Tomasz Korbak, Ethan Perez, and Christopher~L Buckley.
\newblock Rl with kl penalties is better viewed as bayesian inference.
\newblock \emph{arXiv preprint arXiv:2205.11275}, 2022.

\bibitem[Kuhn et~al.(2022)Kuhn, Gal, and Farquhar]{kuhn2022clam}
Lorenz Kuhn, Yarin Gal, and Sebastian Farquhar.
\newblock Clam: Selective clarification for ambiguous questions with large language models.
\newblock \emph{arXiv preprint arXiv:2212.07769}, 2022.

\bibitem[Lakshminarayanan et~al.(2017)Lakshminarayanan, Pritzel, and Blundell]{lakshminarayanan2017simple}
Balaji Lakshminarayanan, Alexander Pritzel, and Charles Blundell.
\newblock Simple and scalable predictive uncertainty estimation using deep ensembles.
\newblock \emph{Advances in neural information processing systems}, 30, 2017.

\bibitem[Lee et~al.(2015)Lee, Purushwalkam, Cogswell, Crandall, and Batra]{lee2015m}
Stefan Lee, Senthil Purushwalkam, Michael Cogswell, David Crandall, and Dhruv Batra.
\newblock Why m heads are better than one: Training a diverse ensemble of deep networks.
\newblock \emph{arXiv preprint arXiv:1511.06314}, 2015.

\bibitem[Li et~al.(2022)Li, Puig, Paxton, Du, Wang, Fan, Chen, Huang, Aky{\"u}rek, Anandkumar, et~al.]{li2022pre}
Shuang Li, Xavier Puig, Chris Paxton, Yilun Du, Clinton Wang, Linxi Fan, Tao Chen, De-An Huang, Ekin Aky{\"u}rek, Anima Anandkumar, et~al.
\newblock Pre-trained language models for interactive decision-making.
\newblock \emph{Advances in Neural Information Processing Systems}, 35:\penalty0 31199--31212, 2022.

\bibitem[Lin et~al.(2022)Lin, Hilton, and Evans]{lin2022teaching}
Stephanie Lin, Jacob Hilton, and Owain Evans.
\newblock Teaching models to express their uncertainty in words.
\newblock \emph{arXiv preprint arXiv:2205.14334}, 2022.

\bibitem[Loshchilov \& Hutter(2019)Loshchilov and Hutter]{loshchilov2018decoupled}
Ilya Loshchilov and Frank Hutter.
\newblock Decoupled weight decay regularization.
\newblock In \emph{International Conference on Learning Representations}, 2019.
\newblock URL \url{https://openreview.net/forum?id=Bkg6RiCqY7}.

\bibitem[Mangrulkar et~al.(2022)Mangrulkar, Gugger, Debut, Belkada, and Paul]{peft}
Sourab Mangrulkar, Sylvain Gugger, Lysandre Debut, Younes Belkada, and Sayak Paul.
\newblock Peft: State-of-the-art parameter-efficient fine-tuning methods.
\newblock \url{https://github.com/huggingface/peft}, 2022.

\bibitem[Mihaylov et~al.(2018)Mihaylov, Clark, Khot, and Sabharwal]{OpenBookQA2018}
Todor Mihaylov, Peter Clark, Tushar Khot, and Ashish Sabharwal.
\newblock Can a suit of armor conduct electricity? a new dataset for open book question answering.
\newblock In \emph{EMNLP}, 2018.

\bibitem[Neal(2012)]{neal2012bayesian}
Radford~M Neal.
\newblock \emph{Bayesian learning for neural networks}, volume 118.
\newblock Springer Science \& Business Media, 2012.

\bibitem[OpenAI(2023)]{OpenAI2023GPT4TR}
OpenAI.
\newblock Gpt-4 technical report.
\newblock \emph{ArXiv}, abs/2303.08774, 2023.
\newblock URL \url{https://api.semanticscholar.org/CorpusID:257532815}.

\bibitem[Ouyang et~al.(2022)Ouyang, Wu, Jiang, Almeida, Wainwright, Mishkin, Zhang, Agarwal, Slama, Ray, et~al.]{ouyang2022training}
Long Ouyang, Jeffrey Wu, Xu~Jiang, Diogo Almeida, Carroll Wainwright, Pamela Mishkin, Chong Zhang, Sandhini Agarwal, Katarina Slama, Alex Ray, et~al.
\newblock Training language models to follow instructions with human feedback.
\newblock \emph{Advances in Neural Information Processing Systems}, 35:\penalty0 27730--27744, 2022.

\bibitem[Ovadia et~al.(2019)Ovadia, Fertig, Ren, Nado, Sculley, Nowozin, Dillon, Lakshminarayanan, and Snoek]{ovadia2019can}
Yaniv Ovadia, Emily Fertig, Jie Ren, Zachary Nado, David Sculley, Sebastian Nowozin, Joshua Dillon, Balaji Lakshminarayanan, and Jasper Snoek.
\newblock Can you trust your model's uncertainty? evaluating predictive uncertainty under dataset shift.
\newblock \emph{Advances in neural information processing systems}, 32, 2019.

\bibitem[Park \& Caragea(2022)Park and Caragea]{park2022calibration}
Seo~Yeon Park and Cornelia Caragea.
\newblock On the calibration of pre-trained language models using mixup guided by area under the margin and saliency.
\newblock \emph{arXiv preprint arXiv:2203.07559}, 2022.

\bibitem[Perez et~al.(2022)Perez, Huang, Song, Cai, Ring, Aslanides, Glaese, McAleese, and Irving]{perez2022red}
Ethan Perez, Saffron Huang, Francis Song, Trevor Cai, Roman Ring, John Aslanides, Amelia Glaese, Nat McAleese, and Geoffrey Irving.
\newblock Red teaming language models with language models.
\newblock \emph{arXiv preprint arXiv:2202.03286}, 2022.

\bibitem[Radford et~al.(2019)Radford, Wu, Child, Luan, Amodei, and Sutskever]{radford2019language}
Alec Radford, Jeff Wu, Rewon Child, David Luan, Dario Amodei, and Ilya Sutskever.
\newblock Language models are unsupervised multitask learners.
\newblock 2019.

\bibitem[Sankararaman et~al.(2022)Sankararaman, Wang, and Fang]{sankararaman2022bayesformer}
Karthik~Abinav Sankararaman, Sinong Wang, and Han Fang.
\newblock Bayesformer: Transformer with uncertainty estimation.
\newblock \emph{arXiv preprint arXiv:2206.00826}, 2022.

\bibitem[Schulman et~al.(2017)Schulman, Wolski, Dhariwal, Radford, and Klimov]{schulman2017proximal}
John Schulman, Filip Wolski, Prafulla Dhariwal, Alec Radford, and Oleg Klimov.
\newblock Proximal policy optimization algorithms.
\newblock \emph{arXiv preprint arXiv:1707.06347}, 2017.

\bibitem[Singhal et~al.(2023)Singhal, Azizi, Tu, Mahdavi, Wei, Chung, Scales, Tanwani, Cole-Lewis, Pfohl, et~al.]{singhal2023large}
Karan Singhal, Shekoofeh Azizi, Tao Tu, S~Sara Mahdavi, Jason Wei, Hyung~Won Chung, Nathan Scales, Ajay Tanwani, Heather Cole-Lewis, Stephen Pfohl, et~al.
\newblock Large language models encode clinical knowledge.
\newblock \emph{Nature}, pp.\  1--9, 2023.

\bibitem[Sun et~al.(2022)Sun, Yan, Abbeel, and Mordatch]{sun2022quantifying}
Meiqi Sun, Wilson Yan, Pieter Abbeel, and Igor Mordatch.
\newblock Quantifying uncertainty in foundation models via ensembles.
\newblock In \emph{NeurIPS 2022 Workshop on Robustness in Sequence Modeling}, 2022.

\bibitem[Talmor et~al.(2019)Talmor, Herzig, Lourie, and Berant]{talmor-etal-2019-commonsenseqa}
Alon Talmor, Jonathan Herzig, Nicholas Lourie, and Jonathan Berant.
\newblock {C}ommonsense{QA}: A question answering challenge targeting commonsense knowledge.
\newblock In \emph{Proceedings of the 2019 Conference of the North {A}merican Chapter of the Association for Computational Linguistics: Human Language Technologies, Volume 1 (Long and Short Papers)}, pp.\  4149--4158, Minneapolis, Minnesota, June 2019. Association for Computational Linguistics.
\newblock \doi{10.18653/v1/N19-1421}.
\newblock URL \url{https://aclanthology.org/N19-1421}.

\bibitem[Touvron et~al.(2023)Touvron, Lavril, Izacard, Martinet, Lachaux, Lacroix, Rozi{\`e}re, Goyal, Hambro, Azhar, et~al.]{touvron2023llama}
Hugo Touvron, Thibaut Lavril, Gautier Izacard, Xavier Martinet, Marie-Anne Lachaux, Timoth{\'e}e Lacroix, Baptiste Rozi{\`e}re, Naman Goyal, Eric Hambro, Faisal Azhar, et~al.
\newblock Llama: Open and efficient foundation language models.
\newblock \emph{arXiv preprint arXiv:2302.13971}, 2023.

\bibitem[Tran et~al.(2022)Tran, Liu, Dusenberry, Phan, Collier, Ren, Han, Wang, Mariet, Hu, et~al.]{tran2022plex}
Dustin Tran, Jeremiah Liu, Michael~W Dusenberry, Du~Phan, Mark Collier, Jie Ren, Kehang Han, Zi~Wang, Zelda Mariet, Huiyi Hu, et~al.
\newblock Plex: Towards reliability using pretrained large model extensions.
\newblock \emph{arXiv preprint arXiv:2207.07411}, 2022.

\bibitem[Vazhentsev et~al.(2022)Vazhentsev, Kuzmin, Shelmanov, Tsvigun, Tsymbalov, Fedyanin, Panov, Panchenko, Gusev, Burtsev, et~al.]{vazhentsev2022uncertainty}
Artem Vazhentsev, Gleb Kuzmin, Artem Shelmanov, Akim Tsvigun, Evgenii Tsymbalov, Kirill Fedyanin, Maxim Panov, Alexander Panchenko, Gleb Gusev, Mikhail Burtsev, et~al.
\newblock Uncertainty estimation of transformer predictions for misclassification detection.
\newblock In \emph{Proceedings of the 60th Annual Meeting of the Association for Computational Linguistics (Volume 1: Long Papers)}, pp.\  8237--8252, 2022.

\bibitem[Wei et~al.(2023)Wei, Haghtalab, and Steinhardt]{wei2023jailbroken}
Alexander Wei, Nika Haghtalab, and Jacob Steinhardt.
\newblock Jailbroken: How does llm safety training fail?
\newblock \emph{arXiv preprint arXiv:2307.02483}, 2023.

\bibitem[Wen et~al.(2020)Wen, Tran, and Ba]{wen2020batchensemble}
Yeming Wen, Dustin Tran, and Jimmy Ba.
\newblock Batchensemble: an alternative approach to efficient ensemble and lifelong learning.
\newblock \emph{arXiv preprint arXiv:2002.06715}, 2020.

\bibitem[Wenzel et~al.(2020{\natexlab{a}})Wenzel, Roth, Veeling, {\'S}wi{\k{a}}tkowski, Tran, Mandt, Snoek, Salimans, Jenatton, and Nowozin]{wenzel2020good}
Florian Wenzel, Kevin Roth, Bastiaan~S Veeling, Jakub {\'S}wi{\k{a}}tkowski, Linh Tran, Stephan Mandt, Jasper Snoek, Tim Salimans, Rodolphe Jenatton, and Sebastian Nowozin.
\newblock How good is the bayes posterior in deep neural networks really?
\newblock \emph{arXiv preprint arXiv:2002.02405}, 2020{\natexlab{a}}.

\bibitem[Wenzel et~al.(2020{\natexlab{b}})Wenzel, Snoek, Tran, and Jenatton]{wenzel2020hyperparameter}
Florian Wenzel, Jasper Snoek, Dustin Tran, and Rodolphe Jenatton.
\newblock Hyperparameter ensembles for robustness and uncertainty quantification.
\newblock \emph{Advances in Neural Information Processing Systems}, 33:\penalty0 6514--6527, 2020{\natexlab{b}}.

\bibitem[Wolf et~al.(2020)Wolf, Debut, Sanh, Chaumond, Delangue, Moi, Cistac, Rault, Louf, Funtowicz, Davison, Shleifer, von Platen, Ma, Jernite, Plu, Xu, Scao, Gugger, Drame, Lhoest, and Rush]{wolf-etal-2020-transformers}
Thomas Wolf, Lysandre Debut, Victor Sanh, Julien Chaumond, Clement Delangue, Anthony Moi, Pierric Cistac, Tim Rault, Rémi Louf, Morgan Funtowicz, Joe Davison, Sam Shleifer, Patrick von Platen, Clara Ma, Yacine Jernite, Julien Plu, Canwen Xu, Teven~Le Scao, Sylvain Gugger, Mariama Drame, Quentin Lhoest, and Alexander~M. Rush.
\newblock Transformers: State-of-the-art natural language processing.
\newblock In \emph{Proceedings of the 2020 Conference on Empirical Methods in Natural Language Processing: System Demonstrations}, pp.\  38--45, Online, October 2020. Association for Computational Linguistics.
\newblock URL \url{https://www.aclweb.org/anthology/2020.emnlp-demos.6}.

\bibitem[Wu et~al.(2020)Wu, Zhang, and Ré]{Wu2020Understanding}
Sen Wu, Hongyang~R. Zhang, and Christopher Ré.
\newblock Understanding and improving information transfer in multi-task learning.
\newblock In \emph{International Conference on Learning Representations}, 2020.
\newblock URL \url{https://openreview.net/forum?id=SylzhkBtDB}.

\bibitem[Yang et~al.(2023)Yang, Liu, and Wang]{yang2023fingpt}
Hongyang Yang, Xiao-Yang Liu, and Christina~Dan Wang.
\newblock Fingpt: Open-source financial large language models.
\newblock \emph{arXiv preprint arXiv:2306.06031}, 2023.

\bibitem[Zaidi et~al.(2020)Zaidi, Zela, Elsken, Holmes, Hutter, and Teh]{zaidi2020neural}
Sheheryar Zaidi, Arber Zela, Thomas Elsken, Chris Holmes, Frank Hutter, and Yee~Whye Teh.
\newblock Neural ensemble search for performant and calibrated predictions.
\newblock \emph{arXiv preprint arXiv:2006.08573}, 2\penalty0 (3), 2020.

\bibitem[Zhang et~al.(2017)Zhang, Cisse, Dauphin, and Lopez-Paz]{zhang2017mixup}
Hongyi Zhang, Moustapha Cisse, Yann~N Dauphin, and David Lopez-Paz.
\newblock mixup: Beyond empirical risk minimization.
\newblock \emph{arXiv preprint arXiv:1710.09412}, 2017.

\bibitem[Zhang et~al.(2020)Zhang, Li, Zhang, Chen, and Wilson]{Zhang2020Cyclical}
Ruqi Zhang, Chunyuan Li, Jianyi Zhang, Changyou Chen, and Andrew~Gordon Wilson.
\newblock Cyclical stochastic gradient mcmc for bayesian deep learning.
\newblock In \emph{International Conference on Learning Representations}, 2020.
\newblock URL \url{https://openreview.net/forum?id=rkeS1RVtPS}.

\bibitem[Zhou et~al.(2023)Zhou, Jurafsky, and Hashimoto]{zhou2023navigating}
Kaitlyn Zhou, Dan Jurafsky, and Tatsunori Hashimoto.
\newblock Navigating the grey area: Expressions of overconfidence and uncertainty in language models.
\newblock \emph{arXiv preprint arXiv:2302.13439}, 2023.

\end{thebibliography}

\newpage
\appendix

\newcommand{\BASE}{LoRA (M=1)\xspace}
\newcommand{\LORAENS}{LoRA Ens. (M=5)\xspace}

\newpage
\section{Additional Implementation Details}
\label{app:details}
We provide a summary of the datasets information in Table.~\ref{table:dataset_info}.

\begin{table}[h]
\small
\centering
\caption{Summary of task setting. For mmlu, we combine the development and validation set as the training set and use the origin test split as the validation set, for the rest datasets, we use the default training set and test splits (validation split for cqa) as the validation set.}
 \begin{tabular}{llllll}
 \toprule
%  \multirow{2}{*}{Estimator} & \multirow{2}{*}{Variance lower bound} & \multirow{2}{*}{$\nabla f$ evals per iteration} & \multicolumn{3}{c}{Wall-clock time per iteration}\\
% \cimidrule{4-6}
  & Task  & Size of training set  & Size of validation set & Number of options  \tabularnewline
 \midrule
 &cqa  & 8741 & 1221  & 5  \\
 &obqa & 4957 & 500 & 4  \\
 &arce & 2249 & 2375 & 4 \\
 &arcc & 1119 & 1172& 4  \\
 &mmlu ss. & 397 & 3077& 4  \\
 &mmlu stem & 411 & 3018& 4 \\
 \bottomrule
\end{tabular}
\label{table:dataset_info}
% \vspace{-20pt}
\end{table}

\section{Ablation study on weight decay regularization}
\label{appendix:weight_decay}

\begin{figure}[ht]
    \centering
    \includegraphics[width=0.95\linewidth]{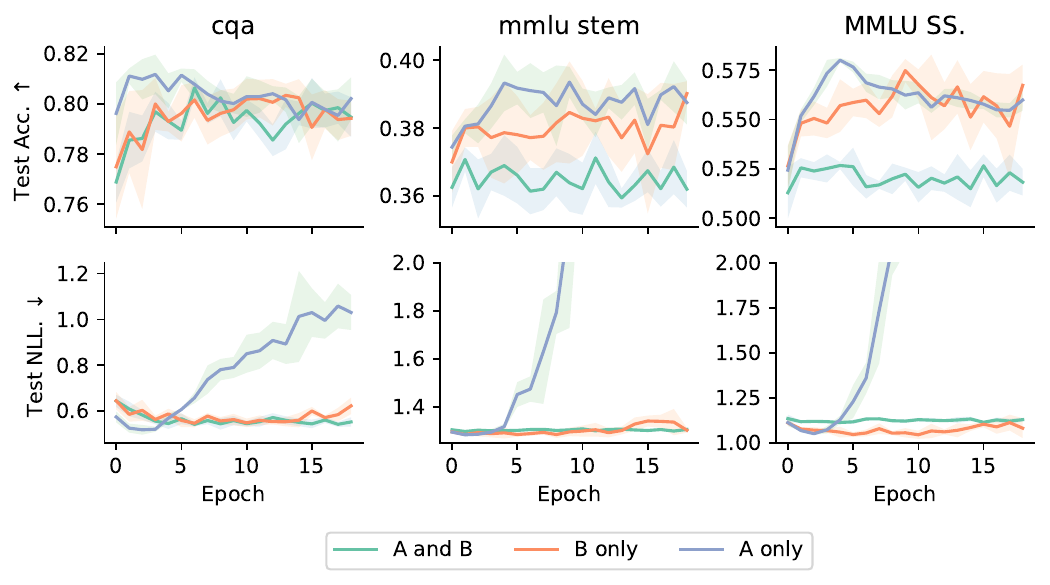}
    \caption{Very large weight decay on A and B underfits the data while regularizing only LoRA A does not resolve overconfidence. Regularizing only LoRA B shows the best performance.}
    \label{fig:wd_ablation}
\end{figure}

Note that LoRA is defined as $\DW = BA$. Therefore we have three ways options for applying weight decay regularization: Regularizing only $A$, regularizing only $B$, or regularizing both. We experiment with these three options on cqa, mmlu ss. and mmlu stem. where we choose $\gamma=1\mathrm{e}{2}$ for cqa and $\gamma=1\mathrm{e}{3}$ for mmlu. We present the performance of these three strategies on validation set in Fig.~\ref{fig:wd_ablation}, where we find that regularizing only $A$ results in high NLL., regularizing both matrices causes significant drops in accuracy, while regularizing only $B$ provides satisfying results in both accuracy and NLL.

\newpage

\newpage
\section{Sample questions}\label{sec:question_sample}
In this section, we provide sample questions from the datasets we experiment with.

\subsection{Commonsense QA (cqa)}
\begin{verbatim}
Q: The sanctions against the school were a punishing blow, 
and they seemed to what the efforts the school had made to change?
Answer Choices:
(a) ignore
(b) enforce
(c) authoritarian
(d) yell at
(e) avoid
A: (a).

Q: Sammy wanted to go to where the people were.  
Where might he go?
Answer Choices:
(a) race track
(b) populated areas
(c) the desert
(d) apartment
(e) roadblock
A: (b).

Q: To locate a choker not located in a jewelry box or 
boutique where would you go?
Answer Choices:
(a) jewelry store
(b) neck
(c) jewlery box
(d) jewelry box
(e) boutique
A: (a).
\end{verbatim}

\subsection{OpenBook QA (obqa)}
\begin{verbatim}
Q: The sun is responsible for
Answer Choices:
(a) puppies learning new tricks
(b) children growing up and getting old
(c) flowers wilting in a vase
(d) plants sprouting, blooming and wilting
A: (d)
Q: When standing miles away from Mount Rushmore
Answer Choices:
(a) the mountains seem very close
(b) the mountains are boring
(c) the mountains look the same as from up close
(d) the mountains seem smaller than in photographs
A: (d)
Q: When food is reduced in the stomach
Answer Choices:
(a) the mind needs time to digest
(b) take a second to digest what I said
(c) nutrients are being deconstructed
(d) reader's digest is a body of works
A: (c)
\end{verbatim}
\subsection{ARC-Easy (arce)}
\begin{verbatim}
Q: Which factor will most likely cause a person to develop
a fever?
Answer Choices:
(a) a leg muscle relaxing after exercise
(b) a bacterial population in the bloodstream
(c) several viral particles on the skin
(d) carbohydrates being digested in the stomach
A: (b)
Q: Lichens are symbiotic organisms made of
green algae and fungi. What do the green algae supply
to the fungi in this symbiotic relationship?
Answer Choices:
(a) carbon dioxide
(b) food
(c) protection
(d) water
A: (b)
Q: When a switch is used in an electrical circuit, the switch can
Answer Choices:
(a) cause the charge to build.
(b) increase and decrease the voltage.
(c) cause the current to change direction.
(d) stop and start the flow of current.
A: (d)
\end{verbatim}
\subsection{ARC-Challenge (arcc)}
\begin{verbatim}
Q: George wants to warm his hands quickly by rubbing them. 
Which skin surface will produce the most heat?
Answer Choices:
(a) dry palms
(b) wet palms
(c) palms covered with oil
(d) palms covered with lotion
A: (a)
Q: Which of the following statements best explains
why magnets usually stick to a refrigerator door?
Answer Choices:
(a) The refrigerator door is smooth.
(b) The refrigerator door contains iron.
(c) The refrigerator door is a good conductor.
(d) The refrigerator door has electric wires in it.
A: (b)
Q: A fold observed in layers of sedimentary rock most
likely resulted from the
Answer Choices:
(a) cooling of flowing magma.
(b) converging of crustal plates.
(c) deposition of river sediments.
(d) solution of carbonate minerals.
A: (b)
\end{verbatim}
\subsection{MMLU social sciences (mmlu ss.)}
\begin{verbatim}
Q: What should a public relations media practitioner do if
she does not know the answer to a reporter's question?
Answer Choices:
(a) Give the reporter other information she is certain is correct.
(b) Say that the information is 'off the record' and will
    be disseminated later.
(c) Say 'I don't know' and promise to provide the
    information later.
(d) Say 'no comment,' rather than appear uninformed.
A: (c).

Q: In issues management, what is the most proactive approach to
addressing negative or misleading information posted online about
your organization?
Answer Choices:
(a) Buy domain names that could be used by opposition groups.
(b) Post anonymous comments on blogs to combat this information.
(c) Prepare a news release that discredits the inaccurate
    information.
(d) Make policy changes to address complaints highlighted on
    these sites.
A: (d).

Q: Which of these statements is true of the Vatican in 2010 at
the time of the accusations of child abuse cover-ups?
Answer Choices:
(a) There was a coordinated media response.
(b) Consistent messages were communicated.
(c) Criticisms were taken as attacks on the Catholic Church.
(d) The credibility of the Vatican was upheld.
A: (c).
\end{verbatim}
\subsection{MMLU STEM (mmlu stem)}
\begin{verbatim}
Q: Let V be the set of all real polynomials p(x). Let
transformations T, S be defined on V by T:p(x) -> xp(x) 
and S:p(x) -> p'(x) = d/dx p(x), and interpret (ST)(p(x)) 
as S(T(p(x))). Which of the following is true?
Answer Choices:
(a) ST = 0
(b) ST = T
(c) ST = TS
(d) ST - TS is the identity map of V onto itself.
A: (d).

Q: A tank initially contains a salt solution of 3 grams
of salt dissolved in 100 liters of water. A salt solution
containing 0.02 grams of salt per liter of water is sprayed into 
the tank at a rate of 4 liters per minute.  The sprayed solution 
is continually mixed with the salt solution in the tank, and the 
mixture flows out of the tank at a rate of 4 liters per minute. 
If the mixing is instantaneous, how many grams of salt are in
the tank after 100 minutes have elapsed?
Answer Choices:
(a) 2
(b) 2 - e^-2
(c) 2 + e^-2
(d) 2 + e^-4
A: (d).

Q: Let A be a real 2x2 matrix. Which of the following statements
must be true?
I. All of the entries of A^2 are nonnegative.
II. The determinant of A^2 is nonnegative.
III. If A has two distinct eigenvalues, then A^2 has two
distinct eigenvalues.
Answer Choices:
(a) I only
(b) II only
(c) III only
(d) II and III only
A: (b).
\end{verbatim}

\end{document}